# A programmable actuator for combined motion and connection and its application to modular robot

Li Zhu, Didier El Baz

LAAS-CNRS, Université de Toulouse, CNRS Toulouse, France

**Abstract**

This paper proposes a new type of actuator at millimeter scale, which is based on Simplified Electro-Permanent (SEP) magnets. The new actuator can achieve connection and smooth motion by controlling the polarity of SEP magnets. Analyses based on numerical simulation are used to design a prototype. A dead-time controllable H-bridge and its multiplex design are proposed for controlling the new actuator and simplifying the electronic circuit. Finally, the new actuator is implemented in the DILI modular reconfigurable robot system. The experimental results show that with this new actuator, the DILI module can move smoothly and connect to other modules without power supply during connection. The maximum speed of DILI module is 20mm/s.

## 1. Introduction

Nowadays, robotic systems have become an indispensable part of the human production process and many human activities have been replaced by robots. In general, these robots are designed for specific-use according to mission requirement and the environment. Researchers have developed Modular Reconfigurable Robot (MRR) systems [1] that are composed of many basic modules with certain functions of motion, connection, perception, and actuation. They can adapt to environment and goals by connecting and disconnecting. For example, reconfigurable conveyors can adapt to changing goals; worm-like robots can also cross a rugged terrain by transforming into quadruped robots, and can form a ring-shaped configuration in a planar environment [2]. Since CEBOT, the first MRR that has been proposed in 1980's [1], more than one hundred MRR systems have been developed. However, most of them are still in the research stage. Miniaturization is one of the main reasons that stop MRR systems to be applied.

MRRs usually contain different systems for motion (actuation) and connection. The motion and connection system are also major obstacles in downsizing modules; typically, they occupy more than 50% of the volume; their weight is also nonnegligible and they consume a lot of energy [3].

Motion systems permit one to move and reconfigure a modular robot. Many attempts to reduce the size and weight of motion system have been made. For instance, the Miche robots at MIT [4] can disengage their magnetic couplings and fall apart under the influence of gravity. In addition to gravity, random movements are also used for modular recombination like in the Pebbles robot [5], and the stochastic 3D robot [6]. However, this kind of MRR system presents drawbacks. For example, they need a long time to reach the desired configuration and have a low success rate for the reconfiguration.

Connection systems ensure stability of a subset of modular robots. Many techniques have been proposed like latches, grippers, hooks. These mechanical systems have several drawbacks: they are difficult to implement and have a nonnegligible size. Permanent magnets can provide a tight connection between modules, but they require manual intervention which is contrary to the concept of autonomous robots. Electro-magnets are easy to control and act fast; however, they require a continuous power supply.

In this paper, we propose a new robotic actuator that can achieve very smooth motion and connection with only one system. Moreover, there is no energy consumption when modules are connected. We also present its application to the DILI MRR system. This work is an extension of the Smart Surface [7-9] and Smart Blocks [10-13] projects.

The rest of the paper is organized as follows. Section 2 presents the new actuator. Section 3 details the numerical simulation of the SEP magnet. Section 4 deals with electronic circuit design. Section 5 presents the application of the new actuator to the DILI modular reconfigurable robot. Section 6 shows the experimental results. Conclusions and future work are given in Section 7.

## 2. Principle of the actuator

### 2.1 Electro-Permanent magnet

Electro-Permanent (EP) magnet [14] is a type of permanent magnet in which the external magnetic field can be switched on or off by a pulse current. The magnet consists of two sections, one is "hard" magnetic material (high coercvity, e.g. NdFeB) another one is "soft" magnetic material (low coercivity, e.g. Alnico5) capped at both ends with a magnetically soft material (e.g. Iron) and wrapped in a coil (see Figure 1). When the magnetically soft and hard materials have the same magnetization, the EP magnet produces an external magnetic field that corresponds to the on configuration. The EP magnet produces no net external field across its poles (off configuration) when the directions of magnetization are reversed. No electrical power is required to maintain the field.

Due to its small power consumption and the strong connectivity，EP magnet is primarily used to provide a force to fasten objects especially at small scale. Robot Pebbles [5], M-Blocks [15] and Ara (Google's modular phone) [16] are good examples of systems using the EP magnets.

We note that NdFeB has a very large coercivity (1000 $kA/m$), while Alnico5 has a relatively small coercivity (48 $kA/m$). However, both have about the same residual induction: 1.28 $T$ and 1.26 $T$, respectively. When pulse currents pass through the copper coil, the only thing that change is the polarity of Alnico5 magnet, the polarity of NdFeB magnet remains the same. Therefore, the magnetic field changes according to the polarity of Alnico5.

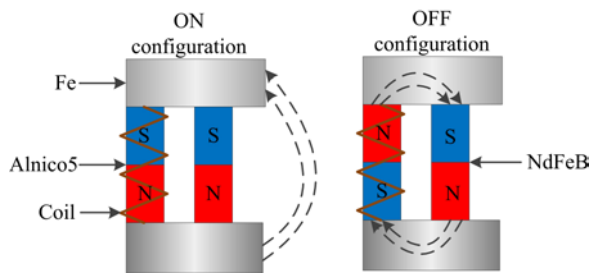

Fig. 1.  The principle of Electro-Permanent magnet

### 2.2 Principle of the Simplified Electro-Permanent magnet

We propose now the principle of a Simplified Electro-Permanent (SEP) magnet for combined smooth motion and connection. To the best of our knowledge, this is the first time that this principle is presented.

SEP magnet relies on Alnico5 wrapped in a copper coil as shown in Figure 2. Polarity of SEP magnet is modified via pulse current. The different parameters of the SEP magnet are detailed in subsection 3.2 where numerical simulations are also presented. The interest in SEP magnet relies on its reduced size and weight that permit one to encompass miniaturization. SEP magnets permit also one to combine motion and connection mechanisms in one system.

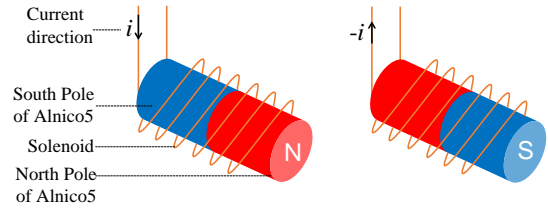

Fig. 2. Principle of Simplified Electro-Permanent magnet

SEP magnet can be used in combination with strong magnets like NdFeB. In a linear motor, one device is fixed, and the other can move. Figure 3 displays a simple example of combined motion and connection mechanism. In this example the SEP magnets are fixed. They are used to move NdFeB magnet with a rectilinear motion. The polarity of SEP magnets changes in order to control the motion of NdFeB. A repulsive force between SEP 1 and NdFeB, and an attractive force between SEP 2 and NdFeB make NdFeB magnet move from left to right (see arrow direction). We note that, when the motion is complete, the NdFeB connects with SEP 2, and no additional power supply is needed. If we want to move the NdFeB in the reverse direction, then we just need to change the polarity of the two SEP magnets.

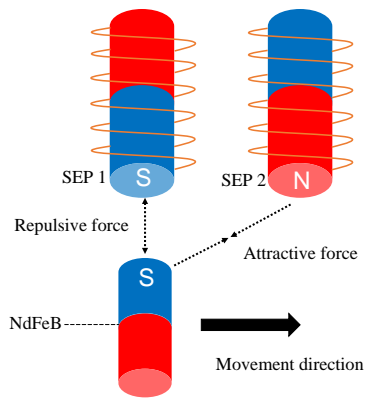

Fig. 3. Motion and connection mechanism

*2.3 Enhanced magnetic field of the new actuator*

Another advantage of SEP magnet is the possibility to have enhanced magnetic field. As a matter of fact, a small continuous electric current in the coil can contribute to increase the magnetic flux density [18]. This enhanced magnetic field can be usefull in some special situation like strengthening the connection. However, since this function consumes power and generates heat, it is only recommended when large connection force, quick movements or complex motions are needed (see subsection 5.2).

**3. Numerical simulation of SEP magnet**

This section aims at validating the design of SEP magnets and finding the most reasonable parameters such as the number of coil turns, coil coverage area, current intensity of pulses.

*3.1 Numerical simulation with COMSOL Multiphysics*

The essence of polarity change of Alnico5 is a succession of magnetization, demagnetization and reverse magnetization that are reflected in the classical hysteresis loop. Our numerical simulation via COMSOL Multiphysics [19] is based on the Jiles-Atherton theory [20]. We consider a two-dimensional symmetric model since the structure of the SEP is axisymmetric.

The model consists of three parts:

- Alnico5 (8 *mm* length and 1.5 *mm* width);
- Copper coil (10 *mm* length and 0.7 *mm* width);
- Surrouding air.

The parameter settings of the numerical simulation are displayed in Table 1.

Table 1 Global parameters for simulation with COMSOL

| Name | Expression | Initial Value | Description |
|---|---|---|---|
| Radius | 1.5[mm] | 0.0015 m | Geometric parameters: the radius of Alnico5 |
| Length | 8[mm] | 0.008 m | Geometric parameters: the length of Alnico5 |
| d_wire_coil | 0.2[mm] | 2E−4 m | Coil parameters: the diameter of the copper wire |
| L_coil | 10[mm] | 0.01 m | Geometric parameters: the length of coil |
| Pulse_Period_p | 0.4[ms] | 4E−4 s | Condition parameters: positive magnetization pulse period |
| P_rise | 0.01[ms] | 1E−5 s | Condition parameters: magnetization pulse rise / fall edge width |
| solver_step | P_rise*0.05 | 5E−7 s | Solver parameters: time of step |
| Pulse_Period_n | 0.4[ms] | 4E−4 s | Condition parameters: negative magnetization pulse time |
| Pulse_hold | 0.4[ms] | 4E−4 s | Condition parameters: positive and negative pulse hold time |
| F_sin | 2[Hz] | 2 Hz | Condition parameters: sinusoidal demagnetization period |
| Br | 0.03[T] | 0.03 T | Material parameters: initial residual magnetic flux density |
| Mr_z | Br/mu0_const | 2.3873E5 A/m | Material parameters: initial residual magnetization |
| Bs | 0.2[T] | 0.2 T | Material parameters: saturated magnetic flux density |
| Ms_z | Bs/mu0_const | 1.5915E6 A/m | Material parameters: saturation magnetization |

Figure 4 displays four important phases when SEP is magnetized and demagnetized thanks to current pulses; we have considered here 20 A pulses. Figure 4 (a) shows the magnetic flux density when it reaches the maximum value. Figure 4 (b) shows the directions of the magnetic field when it starts to reverse. At this moment, the direction of the magnetic field inside Alnico5 is opposite to the direction of the magnetic field outside. The interaction between the magnetic field of the Alnico5 and the one in the coil also causes the nearby magnetic field to become disordered. In Figure 4 (c), the Alnico5 is demagnetized. In Figure 4 (d), the magnetic flux density reaches the maximum value in the opposite direction. The value of the magnetic flux is displayed in color.

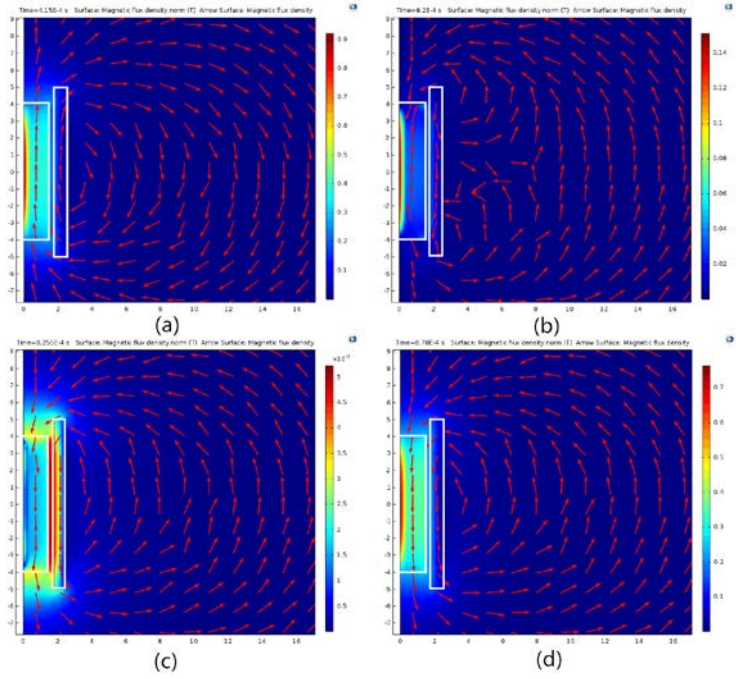

Fig. 4. Magnetic flux density at four representative times with 20 *A* pulses. (a) The maximum value of the flux density; (b) The flux density begins to change direction; (c) Alnico5 is demagnetized; (d) Maximum value of flux density in opposite direction.

*3.2 Simulation results*

Some important results concerning the effect of the number of coil turns, pulse intensity and coverage area of the coil can be obtained via numerical simulation.

3.2.1 Effect of the pulse intensity

As mentioned in section 2, pulse signals are used to change the polarity of SEP magnet. When the pulse signal passes through the copper coil, it generates a magnetic field. In general, the material is fully magnetized when the peak value of the pulse magnetic field reaches 3-5 times the material coercivity [21].

We analyze now the effect of the pulse intensity and consider peaks from 0 *A* to 30 *A*. Figure 5 shows the trends of the magnetic flux density of Alnico5 center, $B_{center}$, and the average magnetic flux density, $B_{average}$. We consider again the case where Alnico5 is totally covered by the coil and the number of coil turns is 250. We observe that the greater the intensity of the pulse, the greater the magnetic flux density.

We note that few energy is needed since the pulse is very short. Thus, it is better to choose an intensity that is enough but also avoids the destruction of the circuits and equipment like 20 *A*.

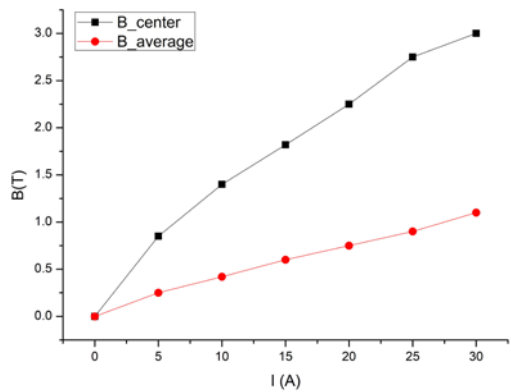

Fig. 5. Effect of the pulse peak on the magnetic flux density

3.2.2 Effect of the number of coil turns

To analyze the effect of the number of coil turns, we make several tests from 0 up to 500 turns and measure the magnetic flux density via our model and numerical simulation with COMSOL. We consider the case where Alnico5 is totally covered by the coil and pulse intensity is 20 A. Figure 6 shows the trends of the magnetic flux density denoted by $B_{center}$ and $B_{average}$. The magnetic flux density is a linear function of the number of coil turns, the more the number of coil turns, the stronger the magnetic field. However, as the number of coil turns increases, the volume of the SEP increases and the SEP also produces more heat. Considering all these limitations, we found that a convenient number of coil turns is situated between 200 to 300 turns.

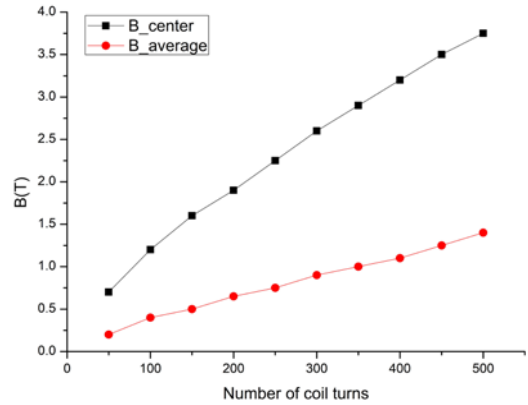

Fig. 6. Effect of number of coil turns on the magnetic flux density

3.2.3 Effect of the coverage area of the coil

We consider now three wrapping configurations for 250 coil turns:

a) Alnico5 is half wrapped, see Figure 7 (a);

b) Alnico5 is exactly wrapped, see Figure 7 (b);

c) Alnico5 is extra wrapped, see Figure 7 (c).

These three configurations correspond to tight, normal and loose wrapping, respectively. Figure 7 displays the distribution of the magnetic flux density according to the coverage area. In Figure 7 (a), the strongest point of magnetic flux density of Alnico5 center is 4 *T*. It can be seen that, if the Alnico5 is not fully covered, then the two ends (in the circle) cannot be magnetized. In Figure 7 (b) and Figure 7 (c), we observe that Alnico5 is totally magnetized and the looser the coverage, the smaller the magnetic flux density of Alnico5 center which are around 2 *T* and 0.9 *T*, respectively. We note that in order to differentiate the effect of the coverage area, the flux densities displayed in Figure 7 do not correspond to peak values; these values are given at 100 microseconds.

Thus, when manufacturing SEP magnet, we better exactly wrap Alnico5 with copper coil.

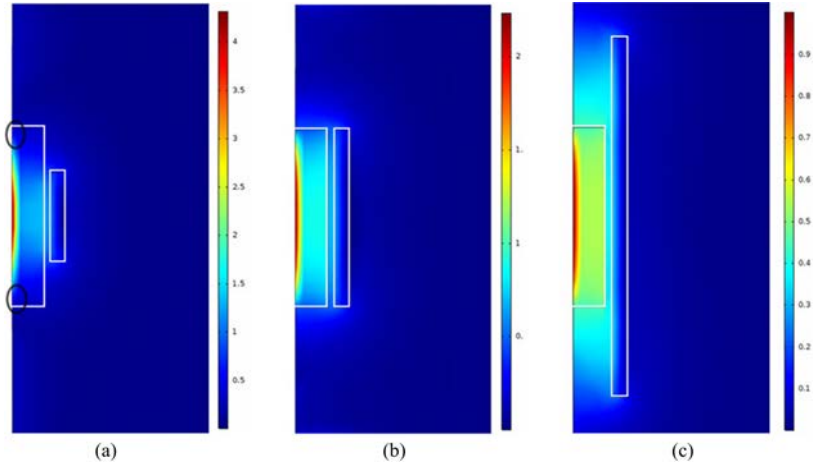

Fig. 7. Effect of the coverage area of the coil. (a) half wrapped; (b) exactly wrapped; (c) extra wrapped

The numerical simulation results play a reference and guidance role for designing SEP. After a comprehensive consideration, we use 250 coil turns with diameter of 0.15 *mm* to make the SEP, and the Alnico5 is exactly wrapped. The pulse intensity is 20 *A*.

## 4. Electronic circuit design

In this section, we present the design of the electronic circuit that we have used in order to control SEP magnet.

*4.1 Dead-time controllable H-bridge*

H-bridge is an electronic circuit that enables a voltage to be applied across a load in either direction. This circuit is commonly used in robotics and other applications to allow DC motors to run forward or backward [22]. MOSFET is often used as the switch of H-bridge since it responds quickly and can pass high current. Nevertheless, there is a dead-time phenomenon in H-bridge built with MOSFET. Dead-time is also called as shoot-through protection or no-overlap Pulse Width Modulation (PWM). The most direct consequences of dead-time are to cause a delay, and the distortion of the output waveform as well as to generate new harmonic components and reduce the effect of active filtering.

We measure the dead-time of H-bridge built with MOSFET; it is equal to 130 *ms*. This delay is a big problem in modular robotics with many actuators. The delay is even longer than the charging time of the capacitor (about 100*ms*), which can cause a very slow motion.

We propose an improved H-bridge so that the dead-time can be controlled, see Figure 8. In the new circuit that we propose, each MOSFET is controlled independently by a specific Input/Output (I/O) of micro-controller, instead of having a half-bridge controlled by only one I/O. The advantage is that the switching time (ON/OFF) of each MOSFET can be precisely adjusted by the microcontroller in function of the expected velocity of modules.

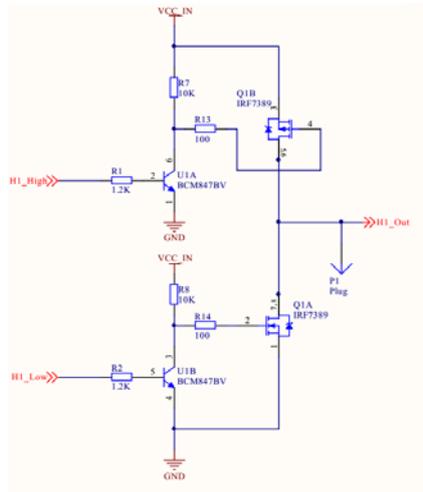

Fig. 8. Improved dead-time controllable Half-bridge

*4.2 Multiplex design*

We also propose a half-bridge multiplex design. With this design, the size and complexity of the circuit can be reduced. The specific structure is shown in Figure 9; the switches S 1 and S 2 form a common semi-H-bridge, like switches S 3 and S 4 or S 5 and S 6 or S 7 and S 8, respectively. Thus, the four semi-H-bridges form a total of three pairs of H-bridges. Figure 9 also presents the main PCB board. Reference is made to [17, chapter V] for more details in multiplex design.

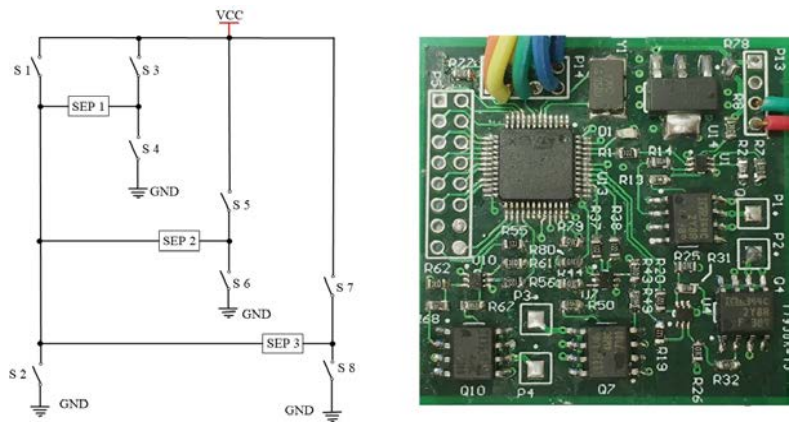

Fig. 9. Half-bridge multiplex design

## 5. Application to DILI modular robot

*5.1 Structure of DILI modular robot*

The new actuator is used in the DIstributed LInear motion modular robot (DILI). Figure 10 presents the full view and cutting view of DILI module; it is a cubic shape robot with side length 1.5 *cm*. The DILI module is shown in Figure 11 with SEP and NdFeB magnets. The basic structure is built via a 3D printer. The total weight of the module including SEP and NdFeB magnet is 12 *grams*. Each module has four work surfaces devoted to both motion and connection. Each module has six big holes for SEP magnets, the diameter is 4 *mm* and the length is 8 *mm*. They penetrate from the outer wall to the inner; the four holes that are close to the wall are embedded inside the wall. This kind of design can not only save space and reduce the weight but is also easy to install. The Alnico5 has a 2.5 *mm* diameter and 8 *mm* length. It is wrapped with 250 turns of 0.15 *mm* enameled copper wire. There are also four small holes for NdFeB magnets, their diameters are 2 *mm* and length are 1.5 *mm*. In order to install NdFeB magnets precisely and keep them perpendicular to the outer wall, we use a non-through design for the small holes. This is very important, if the NdFeB and the wall are not exactly perpendicular, then module motion is not smooth.

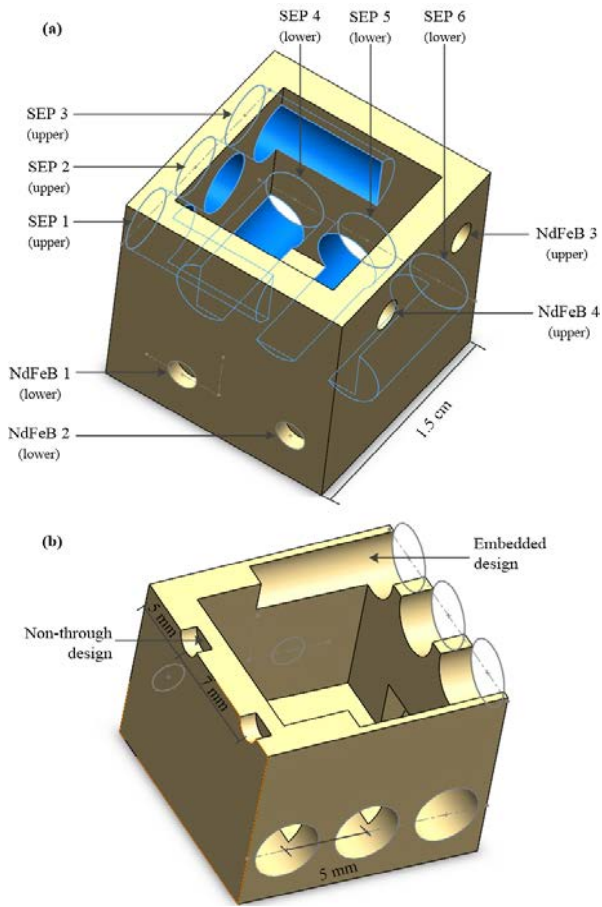

Fig. 10. Structure of DILI robot: (a) full view; (b) cutting view

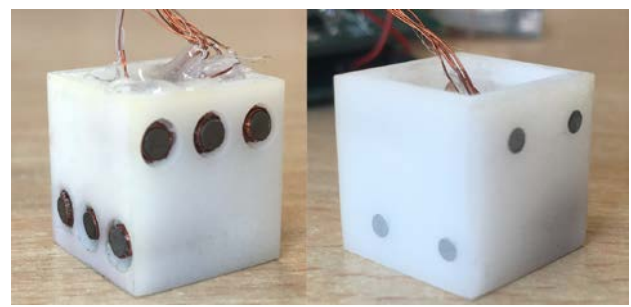

Fig. 11. The outlook of DILI modular reconfigurable robot

The working surfaces of modules have two levels. The surface with NdFeB magnets faces the surface with SEP magnets of another module, as shown in Figure 12. The three SEP magnets of a module and the two NdFeB magnets of another module form a linear motor. Each module can move by itself or can be driven by other modules. Placing the SEP magnets and NdFeB magnets in two layers can not only guarantee every side has the ability of motion, but also makes full use of the space and reduces the volume.

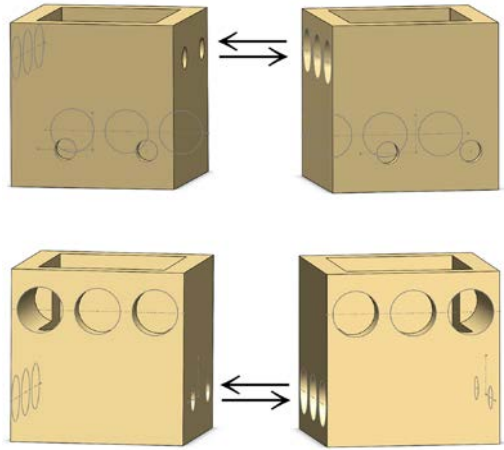

Fig. 12. Relative positions of two DILI modules for working purpose

## 5.2 Motion and connection principle of DILI Robot

DILI module motion relies on the interactions between modules; that is, one module slides along another module. Figure 13 shows the principle of motion and the status of SEP magnets in a complete movement process along a distance of one module (from left to right). Six steps are needed to achieve the whole motion. At each step, only one SEP magnet changes its status and the system only needs energy at that time. The order of the SEP magnets which change the status is 3-1-2-3-1-2. When the motion ends, modules do not need the energy for connection due to the permanent magnetic field generated by Alnico5 and NdFeB magnets. This new actuator can be made as tiny as possible; the limitation is the size of Alnico5 magnet and copper coil. Another advantage of the actuator is that it can move in any direction, due to the controllability of the polarity of Alnico5 magnet. For example, in Figure 13, we can control the module to return to the original location by changing the status of SEP magnets in the following order: 1-3-2-1-3-2.

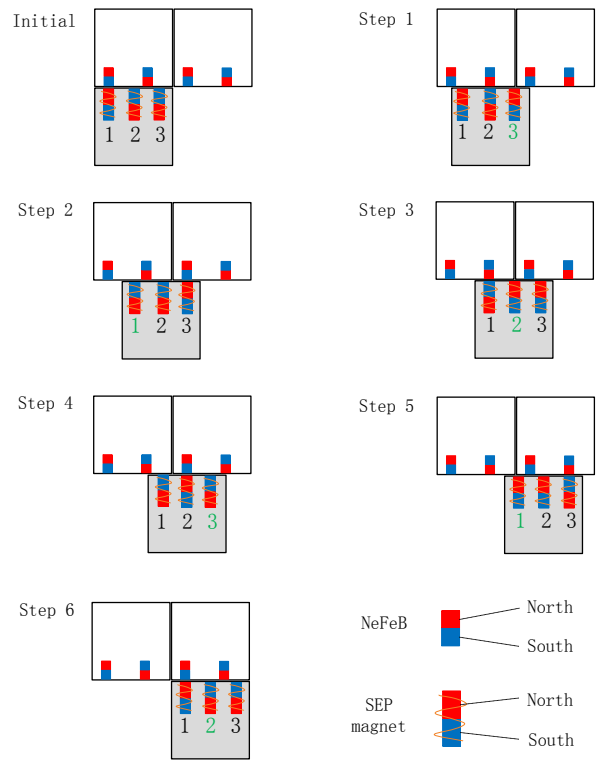

Fig. 13. Status of SEP magnets in one complete movement process along a distance equal to the length of one module

In addition to the simple module motion we have presented above, we can also implement more complex motions like push, pull and carry whereby a connected module is pushed, pulled or carried by a given module. Figure 14 displays examples of complex motions of the DILI robot system.

- Push, shows module 1 pushes module 2 from left to right;
- Pull, shows module 1 pulls module 2 from right to left;
- Carry, shows module 1 carries module 2 to right or left.

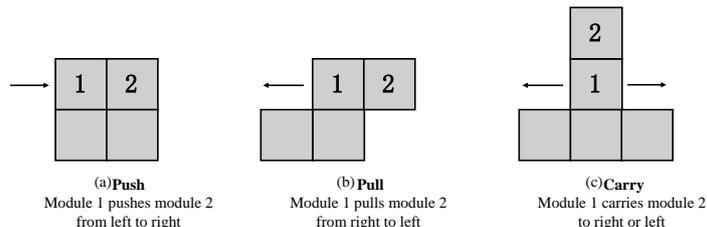

(a) **Push**
Module 1 pushes module 2
from left to right

(b) **Pull**
Module 1 pulls module 2
from right to left

(c) **Carry**
Module 1 carries module 2
to right or left

Fig. 14. Three extended motion capability of DILI module

## 6. Experimental results

In order to validate our design, we have built several DILI modules and a platform to study experimentally the new actuator and DILI module. Since our module can achieve both motion and connection, our experiments focus on these two aspects. For the motion, we study mainly the speed of modules on different materials. For the connection, we study the holding force and the different factors that influence it.

*6.1 Motion and connection principle of DILI Robot*

Pulse generation must be considered carefully when it comes to module motion. A particular attention must be paid to dead-time.

We have designed an ultrasonic velocity tester and tested module speed. Table 2 shows the speed of DILI on different surfaces for three operating modes:

- the stable mode;
- the enhanced mode;
- the fastest mode.

In the stable mode, the DILI module can move without error. The parameters are chosen after many experiments. Dead-time is equal to 40 ms and pulse frequency is 20 Hz. This mode offers reasonable speed as well as smooth motion, as pointed in the video presented in [23]. Experiments are shown on a wood surface. Figure 15 displays photos of the test.

The enhanced mode is relatively similar to the stable mode; the difference is that a 0.8 A continuous current is applied to the coil (for more details, we refer to subsection 2.3). As a consequence, moving modules takes more time due to time spent to generate continuous current.

The fastest mode is essentially achieved by reducing the dead-time to few microseconds; pulse frequency is equal to 200 Hz. It should be noted that, in the fastest mode, the motion may present some error, like an incomplete motion. The reason is that the Alnico5 magnet is not fully magnetized in the fastest mode.

Finally, we note that for a given mode, the speed of a module is almost the same with the different surfaces. The reason is that the motion of DILI is controlled by pulses. Thus, the motion is stepping forward rather than continuous. When a module moves from one position to the next position, the initial speed is zero; the final speed is also zero, there is no speed accumulation. Compared to continuous motion, the step motion is less smooth, therefore, it has a high requirement for the control algorithm.

Table 2 Speed test with different surfaces

| Mode \ Materials | Glass | Paper | Wood | Cement |
|---|---|---|---|---|
| Fastest mode | 20mm/s | 20mm/s | 18mm/s | 18mm/s |
| Stable mode | 13mm/s | 13mm/s | 12mm/s | 12mm/s |
| Enhanced mode | 9mm/s | 9mm/s | 9mm/s | 9mm/s |

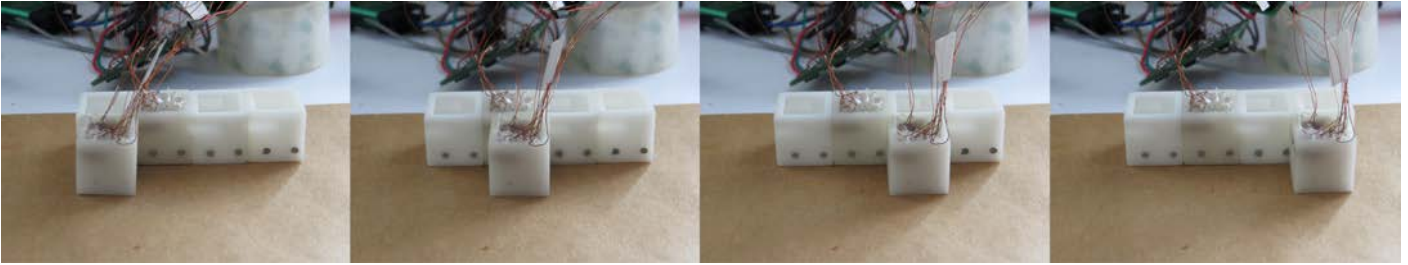

Fig. 15. Speed test of DILI module

*6.2 Holding force test*

We have measured the holding force of DILI modules thanks to a miniature rally meter. The holding force is related to the magnetic field of the SEP magnet. Thus, holding force reaches the maximum when Alnico5 magnet reaches saturation. The intensity of the magnetic field of Alnico5 is related to the number of pulses and the applied voltage. Figure 16 shows the results of holding force test in milli-Newton for different voltage values and different number of pulses. We can summarize the results as follows:

- If the voltage is less than 4 *V*, the circuit cannot produce holding force.
- Holding force increases as the voltage rises.
- Generally, the more the number of pulses, the greater the holding forces. Nevertheless, the holding force meets a maximum.
- The greater the voltage, the less the number of pulses required to reach a maximum holding force. For example, when the voltage is 16 *V*, only one pulse is needed to reach the maximum holding force, 75 *mN*.

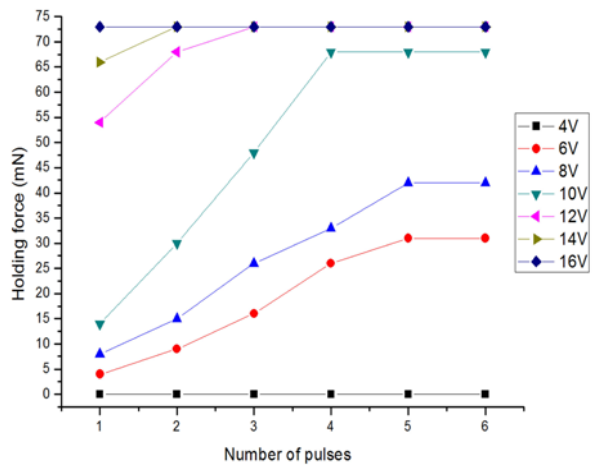

Fig. 16. Holding force test with the pulse signal

We recall that DILI module are used for building a smart conveyor, as a consequence, modules must stay connected after reconfiguration. The smart conveyor must not break up when subject to shakes. The range of holding forces presented in Figure 16 is enough for modules to stay connected when some shakes occur. DILI module weighs only 12 grams. We note that for 30 *mN*, it is hard to separate modules and the connection is very strong when having 75 *mN*.

## 7. Conclusions and future work

In this study, we have proposed a new actuator based on Simplified Electro-Permanent (SEP) magnet. This actuator can achieve both connection and motion at the same time. This actuator provides a miniaturization solution for modular robots; as a consequence, the size, weight and complexity of modular robots are greatly reduced. This new actuator does not require energy consumption while modules are connected, thus, it can save power, which is important for small robots. The design of SEP magnet is based on numerical simulation via COMSOL Multiphysics. Nevertheless, a restriction of the proposed actuator is that the motor is not continuous but step by step.

By using an improved H-bridge, we were able to control dead-time of the actuator. The new actuator is used in the DILI modular reconfigurable robot that is a 12 grams cubic shape robot at centimeter scale. One DILI module contains 6 SEP magnets and 4 NdFeB magnets. DILI module can slide along the surface of other modules with three modes: the enhanced mode, the stable mode, and the fastest mode. The modules have been tested via a series of experiments on many surfaces like wood, paper, glass and cement. In particular, DILI module has a smooth motion when used with stable mode. In that case, the speed of module is around 13 *mm/s*. Modules can also move at the maximum speed of 20 *mm/s* with the fastest mode that corresponds to reduced dead-time.

Future directions of research concern distributed algorithms for controlling DILI modules as well as fault-tolerance and distributed modular robot reconfiguration. Manufacturing will also be carefully considered in conjunction with 3D motion. We note that DILI modules may have a lot of applications. For example, they can be used in smart manufacturing like smart conveyors for drug manufacturing or tiny systems, e.g. clockwork manufacturing. They can be produced for educational purpose or smart robots that evolve on difficult terrain. The concept behind DILI robot may also be used for programmable matter, e.g. furniture, tools and arts.

## Acknowledgments

The authors would like to thank the reviewers for their helpful comments.

Part of this study has been made possible by funding of Chinese Ministry of Education and ANR funding: ANR-2011-BS03-005, Smart Block project.

## Reference

[1] Fukuda T, Nakagawa S. Method of Autonomous Approach, Docking and Detaching Between Cells for Dynamically Reconfigurable Robotic System CEBOT [J]. JSME international journal. Ser. 3, Vibration, control engineering, engineering for industry, 1990, 33(2): 263-268.

[2] Murata S, Kurokawa H. Self-reconfigurable robots [J]. IEEE Robotics & Automation Magazine, 2007, 14(1): 71-78.

[3] White P J, Yim M. Scalable modular self-reconfigurable robots using external actuation [C]. Intelligent Robots and Systems, 2007. IROS 2007. IEEE/RSJ International Conference on. IEEE, 2007: 2773-2778.

[4] Gilpin K, Kotay K, Rus D, et al. Miche: Modular shape formation by self-disassembly [J]. The International Journal of Robotics Research, 2008, 27(3-4): 345-372.

[5] Gilpin K, Knaian A, Rus D. Robot Pebbles: One centimeter modules for programmable matter through self-disassembly [C]. Robotics and Automation (ICRA), 2010 IEEE International Conference on. IEEE, 2010: 2485-2492.

[6] Tolley M T, Lipson H. Fluidic manipulation for scalable stochastic 3D assembly of modular robots [C]. Robotics and Automation (ICRA), 2010 IEEE International Conference on. IEEE, 2010: 2473-2478.

[7] Delettre A, Laurent G J, Le Fort-Piat N. A new contactless conveyor system for handling clean and delicate products using induced air flows [C]. Intelligent Robots and Systems (IROS), 2010 IEEE/RSJ International Conference on. IEEE, 2010: 2351-2356.

[8] El Baz D., Boyer, V., Bourgeois, J., Dedu, E., and Boutoustous, K. Distributed part differentiation in a smart surface [J]. Mechatronics, 2012, 22(5) : 522-530.

[9] Boutoustous K., Laurent G. J., Dedu E., Matignon, L., Bourgeois, J., and Le Fort-Piat, N. Distributed control architecture for smart surfaces [C]. Intelligent Robots and Systems (IROS), 2010 IEEE/RSJ International Conference on. IEEE, 2010: 2018-2024.

[10] Piranda B., Laurent G. J., Bourgeois J., Clévy, C., Möbes, S., and Le Fort-Piat, N. A new concept of planar self-reconfigurable modular robot for conveying microparts [J]. Mechatronics, 2013, 23(7): 906-915.

[11] Mobes S, Laurent G J, Clevy C, et al. Toward a 2D modular and self-reconfigurable robot for conveying microparts[C]//Design, Control and Software Implementation for Distributed MEMS (dMEMS), 2012 Second Workshop on. IEEE, 2012: 7-13.

[12] Laurent G. J., Delettre A., Zeggari R., Yahiaoui, R., Manceau, J. F., and Le Fort-Piat, N. Micropositioning and fast transport using a contactless micro-conveyor [J]. Micromachines, 2014, 5(1): 66-80.

[13] El Baz D, Piranda B, Bourgeois J. A distributed algorithm for a reconfigurable modular surface [C]. Parallel & Distributed Processing Symposium Workshops (IPDPSW), 2014 IEEE International. IEEE, 2014: 1591-1598.

[14] Knaian A N. Electropermanent magnetic connectors and actuators: devices and their application in programmable matter [D]. Massachusetts Institute of Technology, 2010.

[15] Romanishin J W, Gilpin K, Rus D. M-blocks: Momentum-driven, magnetic modular robots [C]. Intelligent Robots and Systems (IROS), 2013 IEEE/RSJ International Conference on. IEEE, 2013: 4288-4295.

[16] http://www.projectara.com

[17] Li Zhu, A distributed modular self-reconfiguring robotic platform based on simplified electro-permanent magnets [D], PhD manuscript, University of Toulouse, February 2018.

[18] Purcell, Edward and Morin, David; Electricity and Magnetism 3rd Edition; Cambridge University Press, New York, 2013. ISBN 978-1-107-01402-2.

[19] Multiphysics C. 5.2, 2015 [J]. COMSOL Multiphysics: a finite element analysis, solver and simulation software for various physics and engineering application, especially coupled phenomena, or multiphysics. URL http://www. comsol. com.

[20] Jiles D. C.，Atherton D. L．Theory of ferromagnetic hysteresis [J]. Journal of Magnetism and Magnetic Materials, 1986, 61(1-2): 48-60..

[21] Lin D, Zhou P, Badics Z, et al. A new nonlinear anisotropic model for soft magnetic materials [J]. IEEE transactions on magnetics, 2006, 42(4): 963-966.

[22] Williams A. Microcontroller projects using the Basic Stamp [M]. Gilroy, CA: CMP Books, 2002.

[23] DILI experiment video, https://youtu.be/kxlJRraiZQI.